%% file: main.tex
\documentclass[letterpaper, 10 pt, journal, twoside]{IEEEtran}

\IEEEoverridecommandlockouts                              

\makeatletter
\let\NAT@parse\undefined
\makeatother
\usepackage[numbers, sort&compress]{natbib}

\usepackage{times}

\usepackage{stackengine}
\usepackage{tikz}
\usepackage{graphicx} 
\usepackage{amsmath} 

\usepackage{makecell}
\usepackage{kotex}

\usepackage{amssymb}  
\usepackage{amsfonts}
\usepackage{subfigure}
\usepackage{ctable}
\usepackage{makecell}

\usepackage[titlenumbered,linesnumbered,ruled,vlined]{algorithm2e}

\usepackage{fancyvrb}
\usepackage{dblfloatfix}
\usepackage{xcolor}
\usepackage{verbatimbox}

\usepackage[labelsep=period]{caption}

\usepackage{multicol}

\usepackage{amsthm}
\theoremstyle{definition}

\usepackage{irap_acronyms}
\usepackage{irap_math}
\usepackage{irap_SIunits}
\usepackage{irap_misc}

\usepackage{soul,color}
\usepackage{verbatim} 
\usepackage{multirow}
\usepackage[bookmarks=true]{hyperref}
\usepackage{cleveref}
\usepackage{src/url_bib_utils}
\usepackage{lipsum}
\usepackage{xcolor}

\let\oldnl\nl
\newcommand{\nonl}{\renewcommand{\nl}{\let\nl\oldnl}}

\begin{document}
	
\title{ 
	(LC)$^2$: LiDAR-Camera Loop Constraints \\ For Cross-Modal Place Recognition
}
\author{Alex Junho Lee${}^{1}$, Seungwon Song${}^{1}$, Hyungtae Lim${}^{2}$, Woojoo Lee${}^{2}$ and Hyun Myung${}^{2*}, $~\IEEEmembership{Senior~Member, IEEE}%
	
	\thanks{Manuscript received: November, 13, 2022; Revised February, 17, 2023; Accepted April, 5, 2023.}
	\thanks{This paper was recommended for publication by Editor Sven Behnke upon evaluation of the Associate Editor and Reviewers' comments.
		This work was supported by Institute of Information \& communications Technology Planning \& Evaluation (IITP) grant funded by the Korea government(MSIT) (No. 2021-0-00230, development of real·virtual environmental analysis based adaptive interaction technology). The students are supported by the BK21 FOUR from the Ministry of Education (Republic of Korea).}
	\thanks{$^{1}$Alex Junho Lee and Seungwon Song are with the Robotics Lab, Research and Development Division, Hyundai Motor Company, Uiwang 16082, South Korea. {\tt\small [alexjunholee, sswan55]@gmail.com}}
	\thanks{$^{2}$H. Lim, W. Lee, and Hyun Myung are with the School of Electrical Engineering, KAIST, Daejeon, Republic of Korea {\tt\small [shapelim, dnwn24, hmyung]@kaist.ac.kr}}
	\thanks{$^{*}$Corresponding author: Prof. Hyun Myung}
	\thanks{Digital Object Identifier (DOI): see top of this page.}
	}

	\markboth{IEEE Robotics and Automation Letters. Preprint Version. Accepted April, 2023}
	{Lee \MakeLowercase{\textit{et al.}}: (LC)$^2$: LiDAR-Camera Loop Constraints For Cross-Modal Place Recognition} 
	\maketitle
	\input{src/abstract}
	
	\IEEEpeerreviewmaketitle
	
	\acresetall
	
	\input{src/introduction.tex}

\input{src/related_work.tex}
	\input{src/method.tex}
	\input{src/experimental_results.tex}
	\input{src/conclusion.tex}

	\renewcommand*{\bibfont}{\small}
	\bibliographystyle{IEEEtranN} 
	\bibliography{string-short,references}
	
	
	\vfill
	
\end{document}

%% file: src/abstract.tex
\begin{abstract}

Localization has been a challenging task for autonomous navigation. A loop detection algorithm must overcome environmental changes for the place recognition and re-localization of robots. Therefore, deep learning has been extensively studied for the consistent transformation of measurements into localization descriptors. Street view images are easily accessible; however, images are vulnerable to appearance changes. LiDAR can robustly provide precise structural information. However, constructing a point cloud database is expensive, and point clouds exist only in limited places. Different from previous works that train networks to produce shared embedding directly between the 2D image and 3D point cloud, we transform both data into 2.5D depth images for matching. In this work, we propose a novel cross-matching method, called \textit{(LC)$^2$}, for achieving LiDAR localization without a prior point cloud map. To this end, LiDAR measurements are expressed in the form of range images before matching them to reduce the modality discrepancy. Subsequently, the network is trained to extract localization descriptors from disparity and range images. Next, the best matches are employed as a loop factor in a pose graph. Using public datasets that include multiple sessions in significantly different lighting conditions, we demonstrated that LiDAR-based navigation systems could be optimized from image databases and vice versa. 

\end{abstract}

%% file: src/introduction.tex
\section{Introduction}
\label{sec:intro}
\IEEEPARstart{G}{lobal} localization is a key problem in mobile robotics. Although the global navigation satellite system (GNSS)
can provide accurate location data in open areas, it may fail to provide correct positions
in urban or indoor environments owing to occlusion or blackout~\cite{kos2010effects}. Thus,
mobile robots should adopt localization systems to determine their positions on a map based
on observations from an operational sensor system. Among them, visual sensors
such as cameras, are widely used for their price competency and data intuitiveness. However,
cameras suffer from appearance changes and require algorithms to be robust to such
variances. Focusing on robust visual landmarks for place recognition, methods such
as bag-of-words model~\cite{GalvezTRO12} and approaches using convolutional neural networks
(CNNs)~\cite{torii201524, babenko2015aggregating, arandjelovic2016netvlad, leyvavallina2021gcl}
have been introduced.

\input{src/figtex/fig_main.tex}

Despite the advances in image-based place recognition algorithms, cameras are not always the best
option for robot localization because the images are vulnerable to environmental changes.
Therefore, LiDAR-based navigation systems are generally utilized for localization and mapping of mobile robots.
LiDAR-based simultaneous localization and mapping (SLAM) has succeeded in constructing precise point cloud maps~\cite{lin2019loam_livox}
and estimating relative poses at loop closures~\cite{kim2018scan, lim2022quatro}. 
With the provided point clouds obtained from a prior visit, a loop closure can be defined between the current frame and point cloud database. 

However, the point cloud database may not always be available owing to the limited accessibility of LiDAR sensors.
Compared with cameras, LiDAR sensors are bulky and expensive and consume more energy.
Therefore, the existing databases are often captured using vision-based systems owing to their economic feasibility for database construction and update.
However, the visual information and a point cloud differ in data representation, known as the discrepancy of modality. Thus these abundant visual databases cannot be directly used for LiDAR-based platforms.
Therefore, it is necessary to find methods that enable the utilization of these visual databases for LiDAR based systems.

To address this problem, image-to-point-cloud fusion has been studied recently.
Many researchers have studied methods for extracting structural information from visual data to align with the
point cloud~\cite{caselitz2016monocular, kim2018stereo, feng20192d3d, yu2020monocular, wang2021p2}.
Other studies have proposed directly matching visual data with point cloud using deep neural networks (DNNs)~\cite{cattaneo2019cmrnet, chang2021hypermap, li2021deepi2p}.
These studies demonstrated some possibilities that allowed vision-based systems to operate within known point cloud maps~(i.e. 2D-to-3D matching). However, these methods are not appropriate for achieving place recognition using a point
cloud as a query on the visual databases~(i.e. 3D-to-2D matching). Additionally, 3D-to-2D matching is more difficult because the precision of structural details extracted from the images is insufficient
to directly build a point cloud map for LiDAR-based place recognition. Hence, 3D-to-2D matching has not yet been adequately examined.

Therefore, we propose a novel method, called \textit{(LC)$^2$}, to achieve 3D-to-2D matching for LiDAR-based systems. To this end,
we formulate this problem as depth image matching by transforming both the image and point cloud data into a depth form.
Not to limit the application within scale-aware depth obtained from SLAM, we created a database of unscaled depths from images and LiDAR scans, and trained a neural network to encode depth into
localization descriptors. Further, with the relative poses between the LiDAR scan and the geotags of images, a loop closure is composed without a point cloud database, as illustrated in Fig.~\ref{fig:cover}.

The main contributions of this study are as follows:

\begin{itemize}
	\item We propose a vision-based place recognition pipeline for LiDAR-based navigation systems to enable LiDAR localization without a point cloud database.
	\item Our module provides a shared embedding between the point clouds and images along with cross-modal loop closures, which could be formulated as global pose constraints for both vision and depth-based systems.
    \item We evaluate our system using public datasets with broad environmental changes and verify that the proposed LiDAR-to-camera matching is robust to appearance changes.
\end{itemize}

%% file: src/figtex/fig_main.tex
\begin{figure}[!t]
	\centering
	\vspace{-6mm}
	\captionsetup{font=footnotesize}
	\includegraphics[width=0.85\columnwidth]{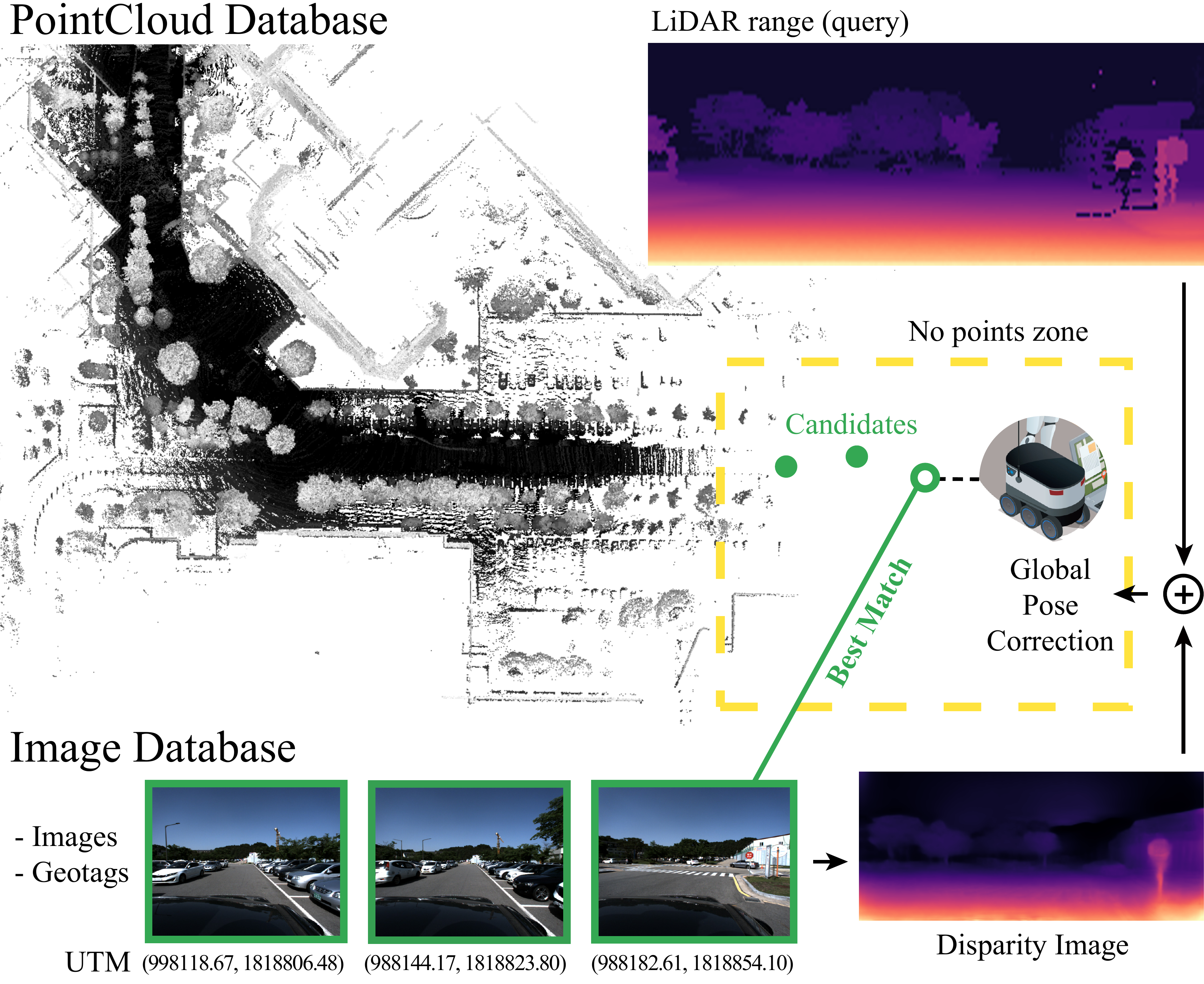}
	\caption
	{
	 	Example of our cross-modal matching scenario to overcome the database scarcity of point clouds.
	 	For instance, when a LiDAR-based SLAM system traverses through an area without prior point clouds (yellow dash),
	 	we propose correcting the global poses with the geotag candidates (green) from the image database
	 	and re-localizing the LiDAR-based system.
	}
	\vspace{-4mm}
	\label{fig:cover}
\end{figure}

%% file: src/related_work.tex
\input{src/figtex/fig_outline.tex}

\section{Related Works}
\label{sec:related}
In this section, we first review place recognition and localization methods based on visual and depth
features and then introduce the multi-modal fusion methods.

\subsection{Image-Based Place Recognition}
\label{sec:vis_loc}
Visual place recognition has been developed based on bag-of-words~\cite{GalvezTRO12} or view synthesis~\cite{torii201524},
which primarily use handcrafted features or their vector of locally aggregated descriptors (VLAD)~\cite{jegou2011aggregating} and is less robust to noise.
To improve the generalization ability, CNN models have been proposed to transform the image features into
localization descriptors~\cite{babenko2015aggregating}. This approach has exhibited superior robustness to changes in appearance.
To efficiently translate place-distinctive image encoding into localization descriptors, the negative and positive samples were trained in pairs using weak supervision in NetVLAD~\cite{arandjelovic2016netvlad} with a triplet loss~\cite{hermans2017defense}. Radenovic \textit{et al.} discovered more consistent and distinctive feature
representation from images, such as contrastive loss with generalized mean (GeM) pooling~\cite{radenovic2018fine}.
The study considers every image with sufficiently large number of co-observed 3D points
or similar features for a training pair.
This differs from weak supervision with triplet loss, which selects pairs by the image location and their descriptor distance. 
In~\cite{garg2019look}, a method to improve feature matching with depth estimation has been introduced, opening a potential
to enhance place recognition with estimated depth. In recent studies, methods unifying local and global features have been
proposed for expansion over large-scale visual place recognition~\cite{cao2020unifying} to geometrically
verify that the local feature matches after global image searching. 

\subsection{Point-Cloud-Based Place Recognition}
\label{sec:depth_loc}
LiDARs are widely used in robotics owing to their high spatial resolution. However, the rich textures in RGB images are rarely recorded in LiDAR and only structural information can be used for place recognition. As a result, point cloud-based localization is reduced to a problem of efficiently transforming the geometrical information between points. PointNetVLAD~\cite{uy2018pointnetvlad} was proposed to adopt PointNet~\cite{qi2017pointnet} as a module to transform input point clouds into localization descriptors. The study employed triplet and quadruplet losses to train a discriminative network. The point cloud submaps were cropped and downsampled to a 25~$\times$~25~m bounding box and then transformed to a descriptor by the network. Point cloud representation is spatially sparse and an appropriate downsampling approach must be identified. A method of outlier rejection by a robust kernel with a reduced degree of freedom~\cite{lim2022quatro} was proposed to improve voxel downsampling used in PointNetVLAD. Previous studies have focused on efficient and robust point cloud representation.
However, a projective depth image representation may be sufficient for place recognition tasks that mainly consider on-sight objects. To exploit range images with a dense form, range-image-based classification~\cite{meyer2019lasernet, milioto2019rangenet} or Monte Carlo localization~\cite{chen2021range} based on LiDAR-generated range images was studied. Because our goal is to match point cloud to RGB images, we focus on the LiDAR-generated projective range image to match with the natively projective images.

\subsection{Cross-Modal Place Recognition}
\label{sec:multi_loc}
Cross-modal matching, particularly camera-LiDAR fusion, is required in two scenarios. The first is to enhance the less accurate sensor using more sophisticated sensors and the second is to overcome the absence of references with a less accurate but easily accessible database. In the early stages of image-to-point-cloud fusion, LiDAR points were augmented with intensity values and rendered for further matching~\cite{wolcott2014visual}. However, the process of photometric rendering and matching is computationally expensive. Therefore, subsequent studies focused on extracting shared representation between modalities. In \cite{caselitz2016monocular}, the researchers used the 3D locations of visual features with a triangulated depth to construct local feature clouds and aligned them with the LiDAR map via graph optimization. Similarly, in \cite{kim2018stereo}, the dense depth from a stereo camera was used to align the images and LiDAR data. In the study, relative poses were calculated by minimizing the sum of the depth residual between the sensor point clouds. To use higher-level features, Yu \textit{et al.} aligned the images and point clouds with co-visible 2D to 3D edge constraints~\cite{yu2020monocular}.

For a generalization with deep learning-based approaches, 2D3D-MatchNet~\cite{feng20192d3d} used a network to cross-match the features in both the images and ground-removed point cloud submaps. In the P2-Net~\cite{wang2021p2}, a batch-hard detector~\cite{mishchuk2017working} was selected to produce shared embedding between the different sensor modalities. However, training and testing were performed only in submaps with sizes of less than a meter, thus this approach remains unsuitable for robotic applications. For large-scale place recognition, global 2D pose was provided using a satellite image and the radar matching network~\cite{tang2020rsl}, or by satellite image and building outline matching~\cite{choi2020brm}. Additionally, the possibilities of camera-LiDAR shared embedding were presented in~\cite{ratz2020oneshot} and~\cite{cattaneo2020global} by transforming a pair of the image and point cloud with CNNs or by transforming the point clouds and images using 3D and 2D CNN, respectively.

Similar to the approaches that establish a shared embedding between the image and point cloud, we propose the transformation of images and point clouds into localization descriptors. However, unlike the previous approaches that directly match the Cartesian sparse point cloud to the dense projective images, we propose to avoid this modality discrepancy by transforming both data into a projected depth form. Subsequently, with the depth images from the image and point cloud, respectively, we search for the best matching images in the database and utilize the best match as a loop constraint. Moreover, we propose to use the location of matched images as a global loop closure factor with pose graph optimization considering that the geotags of images are often available in GPS-denied regions~\cite{xiao2016survey, obeidat2021review}.

%% file: src/figtex/fig_outline.tex
\begin{figure*}[!b]
  \centering
  \captionsetup{font=footnotesize}
  \vspace{-5mm}
  \includegraphics[width=0.85\textwidth]{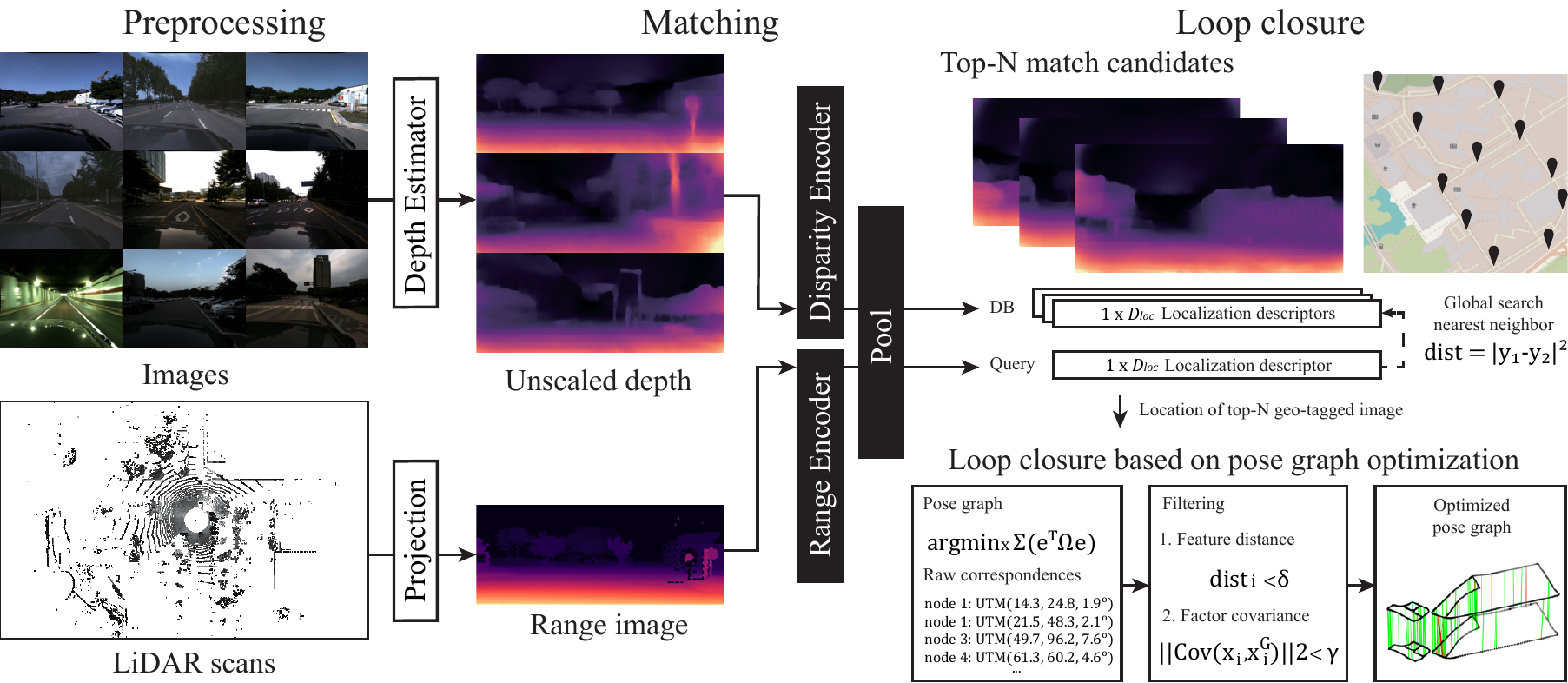}
  \caption
  {
	Pipeline of our LiDAR scan matching with an image database. We first transform the image database into unscaled depth images and LiDAR scan to a range image. Either the range and unscaled depth (i.e. disparity) is used as input for the encoders above. After the image descriptors are delivered from the network, we run a global search (dashed arrow) and pose graph optimization based filtering to ignore the false-positive loops.
  }
  \label{fig:outline}
\end{figure*}

%% file: src/method.tex
\section{(LC)$^2$: Cross-Modal Place Recognition}
\label{sec:method}
In this section, we describe the details of our LiDAR place recognition with an image database, as illustrated
in Fig.~\ref{fig:outline}. To fuse information from different modalities, we first transform the
images and point clouds into the same domain to obtain the depth images. We use a single image for depth estimation; however, sequences of images may also be selected for better performance.
We then train a network to learn the descriptors from each image. Finally, a LiDAR pose graph is optimized with visual loop constraints by running a pose graph optimization based on the match candidates.

\subsection{Data Preprocessing}
\subsubsection{Range Image Generation}
\label{sec:rangeimage}
LiDAR point clouds are provided in a sparse form, with Cartesian coordinates~($x, y, z$). To reduce the expressive differences of different modalities, we represent a point cloud as a range image, $I(u,v)$, whose size is $H \times W$; $H$ and $W$ denote the number of ray channels at its elevation angle and the number of pulses for every single channel along the horizontal direction, respectively. Each pixel,~$(u, v)$, is then assigned to the corresponding point~$p(x,y,z)$ by the following relation:
\begin{align}
\begin{split}
&u = \frac{W}{2}\cdot (1 - \tan^{-1}(y, x) \cdot \pi^{-1}) \\
&v = H \cdot ( F_\text{up} - \sin^{-1}(z \cdot d^{-1})) \cdot F^{-1},
\end{split}
\end{align}
\noindent where $F_\text{up} > 0$ and $F > 0$ denote the upward and total vertical field of view~(FoV), respectively, and $d = \sqrt{x^2 + y^2}$ denotes the depth value. The range image is then resized to the appropriate size for the network input.

\subsubsection{Depth Image Generation}

Because the images captured by camera sensors do not directly contain the distance, depth estimation must be performed to generate depth images. Among the various depth estimation methodologies~\cite{godard2019digging, watson2021temporal}, we exploit ManyDepth~\cite{watson2021temporal} to create depth images because it allows the module to extract depths from both single frame and short sequences. These depth images are fed into the network in a disparity form. Each depth value follows the inverse depth parametrization to make closer objects numerically dominant.

\subsection{Depth-Image-Based Matching}
\subsubsection{Degree of Similarity}
Our proposed pipeline requires sets of geotagged range and monocular images for training. However, although the geo-tagged data pairs are provided, we still need to verify whether the range and corresponding images are sufficiently overlapped. Otherwise, waste pairs whose viewpoints are not overlapped can be regarded as inputs, hindering the training convergence. In particular, as shown in Fig.~\ref{fig:overlap}, the vertical FoV of a typical LiDAR is vertically smaller than that of a camera; it is extremely important to check the extent to which the two areas are overlapped.

\input{src/figtex/fig_overlap.tex}

To resolve this overlap determination problem, the degree of similarity~\cite{leyvavallina2021gcl} is defined to quantitatively approximate the overlapped field of interest between two sensors, as shown in \figref{fig:similarity}. A constant value is assigned to the similarity of the range-disparity pairs measured simultaneously because the extrinsic parameters do not change during the experiment. The overlap measure is calculated with ground truth poses for the pairs defined over different measurements. This degree of similarity $\psi~\in~[0,1]$ is used for weighting the distance between the localization descriptors.

\input{src/figtex/fig_similarity.tex}

\subsubsection{Network Architecture}
We deploy a series of encoders based on VGG16~\cite{simonyan2014very} and a pooling layer.
Given the two types of depth images, these two inputs $x$ are transformed into the feature representations
$f(x)~\in~\mathbb{R}^{H_e\times W_e \times D}$, by the Siamese network architecture~(Fig.~\ref{fig:outline});
$H_e$ and $W_e$ denote the sizes of the outputs from the last layer of encoders, and $D$ represents the dimension of the feature space. 
It should be noted that these two inputs, $x$, are range images from the LiDAR and an unscaled depth from the camera; however, the inputs, $x$,
can be freely selected between the range or disparity (see Section~\ref{sec:result}). The dimension of $f(x)$ is identical to those of the outputs of the last
convolutional layer of VGG16, which is \texttt{conv5}. The features $f(x)$ are then transformed into
the localization descriptor $\hat{f}(x)~\in~\mathbb{R}^{D_{loc}}$ by the pooling layer, where $D_{loc}$ is the dimension of the localization descriptor.

To achieve cross-modality learning, two training phases are proposed by changing the pooling layer. First, we train the network with GeM~\cite{radenovic2018fine} as a pooling layer to generate a similar descriptor when the data is actually measured in a similar place, even if this architecture introduces inputs of the different modalities. Next, the pooling layer is changed to NetVLAD~\cite{arandjelovic2016netvlad} in Phase 2, which forces the encoder output to converge to the appropriate place-distinctive features. This is detailed as follows.

\subsubsection{Phase 1 Using Contrastive Loss}
Empirically, it was found that single-phase learning or weight sharing failed to converge owing to the large modality differences between the two sensors, and the difference in noise characteristics between them caused the divergence of the Siamese network architecture. Therefore, we need to pre-train the two distinct encoders from scratch with the depth images. We aim to create place-distinctive descriptors from scratch, despite the modality discrepancy.

To this end, we use self-supervised pre-training, which has been demonstrated to be highly effective for initialization~\cite{cole2022does}. We propose a modified contrastive loss that can successfully pre-train the encoders. First, we search every overlap ($\psi \neq 0$) between the measurements in the training set and select them in pairs~($i, j$). During Phase 1, range images from the LiDAR scans are horizontally cropped by a fixed size, to increase the number of samples. Details are in Section~\ref{sec:cropaug}. Not only pairs from identical image sources but also cross-sensor pairs are selected to enforce the network to learn the consistent representation over the modality. Subsequently, with the degree of similarity calculated for each pair and a predefined constant $\tau$, the proposed loss enforces the encoders to learn common representations between the overlapping depth images for each pair $(i, j)$:
\begin{align}
	\begin{split}
		\mathcal{L}^{\text{M}}_{i,j} &= \psi_{i,j} \cdot d(x_i , x_j)^2 +\\	&(1-\psi_{i,j}) \cdot \text{max}(\tau - d(x_i , x_j), 0)^2 ,
	\end{split}
\end{align}
where $x_i$ and $x_j$ denote the two input images, $d$ is the distance in the localization descriptor defined by
$d(x_i, x_j) = ||\hat{f}(x_i) -\hat{f}(x_j)||$, and $\hat{f}$ is the output from the pooling layer, as defined in the
previous section. The similarity between the pairs $(i,j)$ is represented as $\psi_{i,j}$ and it motivates the localization
descriptors to have a distance equal to the degree of similarity. The overall loss is then defined by
the summation over all the overlapping pairs ($i,j$): $\mathcal{L}^{\text{M}}~=~\sum_{(i,j)}\mathcal{L}^{\text{M}}_{i,j}$.

\subsubsection{Phase 2 Using Triplet Loss}
After the convergence of contrastive training, we change the pooling layer to NetVLAD and apply triplet margin loss to force the network to reweigh the importance of the depth features for the place recognition task. In this stage, the range images from a LiDAR are cropped by the size of the camera FoV to ensure that the descriptors converge. The triplet samples are selected based on the geometrical distances between the features. For example, positive pairs $\mathbf{p}$ are selected from the samples within 10 meters, whereas the negative pairs $\mathbf{n}$ are randomly sampled from measurements farther than 25 meters. The triplet margin loss consists of the summation over every triplet $k$ defined for a sample $x_i$, as follows:
\begin{align}
	\begin{split}
		\mathcal{L}_{i} = \sum\limits_{(i,k)} l\Bigl( d(x_i, \mathbf{p}_{i,k}) - d(x_i, \mathbf{n}_{i,k}) + m \Bigl),
	\end{split}
\end{align}
where $l$ is the hinge loss $l(x) = \textrm{max}(x, 0)$ and $m$ is a margin. By applying this triplet loss for every
sample $i$ and its selected pair $j$, we train the network to learn place-distinctive localization descriptors from the depth images.

\subsection{Data Augmentation}
We use the sequences of Vision for Visibility Dataset~\cite{lee2022vivid} for training. The depth estimation module
is fine-tuned with \textit{driving-vision} sequences and all the other experiments are conducted using
\textit{driving-full} sequences. All the \textit{day} sequences except \textit{day2} are used for the
Phase~2, and tests are conducted with the \textit{evening} and \textit{night} sequences. The geotags
in the image are assigned by the GNSS signal obtained at the time of image acquisition.

\subsubsection{Range Image Augmentation}
\label{sec:cropaug}
To define the similarity metric between the LiDAR range images and increase the number of samples, we divide the panoramic range image into eight FoV-masked overlapping images, as in~\cite{zamir2010accurate}, and calculate the degree of similarity for training Phase 1.

\subsubsection{Scale Augmentation}
The network is trained to minimize the scale estimation error from the monocular depth estimates, by directly multiplying a random constant to the depth. For training Phase 1, the multiplied random variable modifies up to $r$\% of each depth image, as shown in Fig.~\ref{fig:aug}.

\input{src/figtex/fig_normalization.tex}

\subsection{Loop Closure as a Factor in Graph}
\label{sec:loop_method}
As a result of the training discussed earlier, the place recognition results are obtained as the $N$ closest indices of the database for each query image. However, the raw correspondences are not always correct and must be filtered for trajectory optimization. Because the false-positive loop constraints can result in SLAM divergence, we propose filtering the false-positive loops using pose graph optimization. Assuming sufficient inliers, we construct a pose graph based on the odometry constraints from the LiDAR odometry and raw loop closures from our cross-matching place recognition module. In the factor graph $F$, consisted of nodes $\phi$, the keyframes from odometry $\phi_i$ and geo-tag locations $\phi^G_i$ from each best-matching data are set up as factor nodes and their locations $X_i$, $X^G_i$ are arranged as variable nodes. Our goal is to identify informative loop closure edges $e^G_{i}$ defined between $\phi_i$ and $\phi^G_i$ by solving the maximum a posteriori problem and calculating the information of each loop closure. We assign the initial locations of $X_j$ in a local coordinate and iteratively solve the optimum of the following equation:
\begin{align}
	\begin{split}
		\phi(X) = \prod_{i} \phi_i(X_i).
	\end{split}
\end{align}
The covariances of loops are set higher than the odometry covariances; for example, we set up the odometry covariance as $10^{\text- 2}$ and loop covariance as $10^{\text 4}$. (4) is solved using a Levenberg-Marquardt optimizer and we obtain both variables after optimizing $x_j$ with the information $I^G_{i}$ of each edge $e^G_{i}$. We then filter the false-positive loop closures by the $L2$-norm of the diagonal elements in $I^G_{i}$.


%% file: src/figtex/fig_overlap.tex
\begin{figure}[!t]
	\centering
	\captionsetup{font=footnotesize}
	\includegraphics[width=\columnwidth]{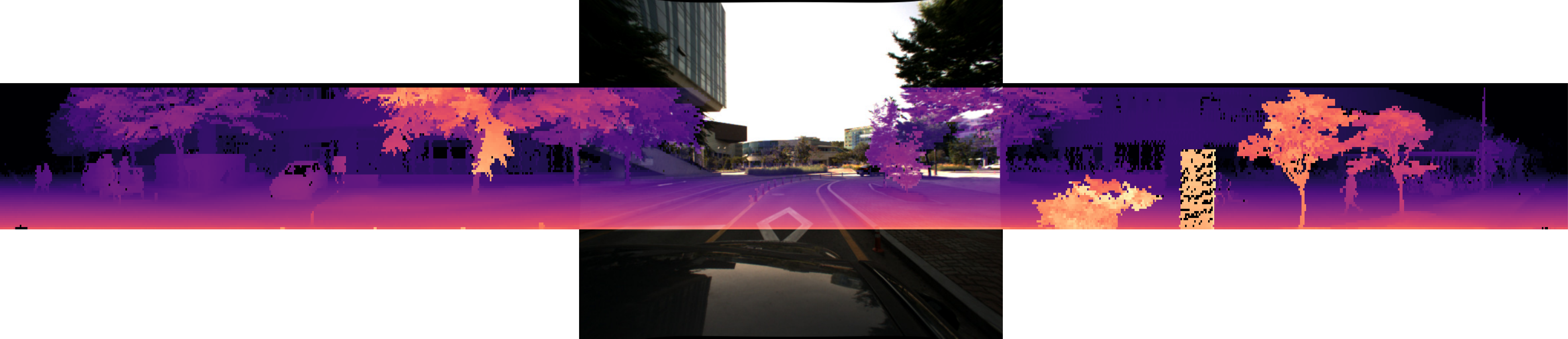}
	\caption
	{
		Sample overlay of RGB and range images from a camera and single LiDAR scan (360$^\circ$). As illustrated above, only small fractions of sensor measurements overlap.
	}
	\label{fig:overlap}
	\vspace{-4mm}
\end{figure}

%% file: src/figtex/fig_similarity.tex
\begin{figure}[!t]
	\centering
	\captionsetup{font=footnotesize}
	\includegraphics[width=\columnwidth]{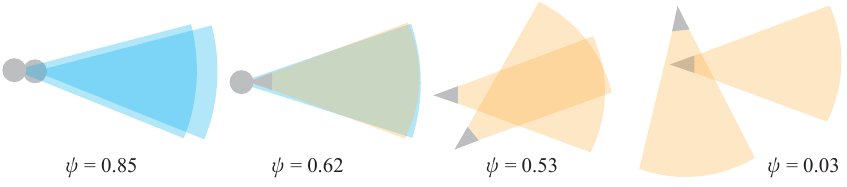}
	\caption
	{
		Visualization of the degree of similarity based on FoV overlap. Here, the LiDAR's (circle) interest area is marked in cyan and the camera's (triangle) in yellow. FoV and maximum effective range were used to define the interest area. The value of $\psi$ is displayed that represents the interest area overlap between 0 (completely distinct) and 1 (completely overlapped).
	}
	\vspace{-7mm}
	\label{fig:similarity}
\end{figure}

%% file: src/figtex/fig_normalization.tex
\begin{figure}[!t]
	\centering
	\captionsetup{font=footnotesize}
	\includegraphics[width=\columnwidth]{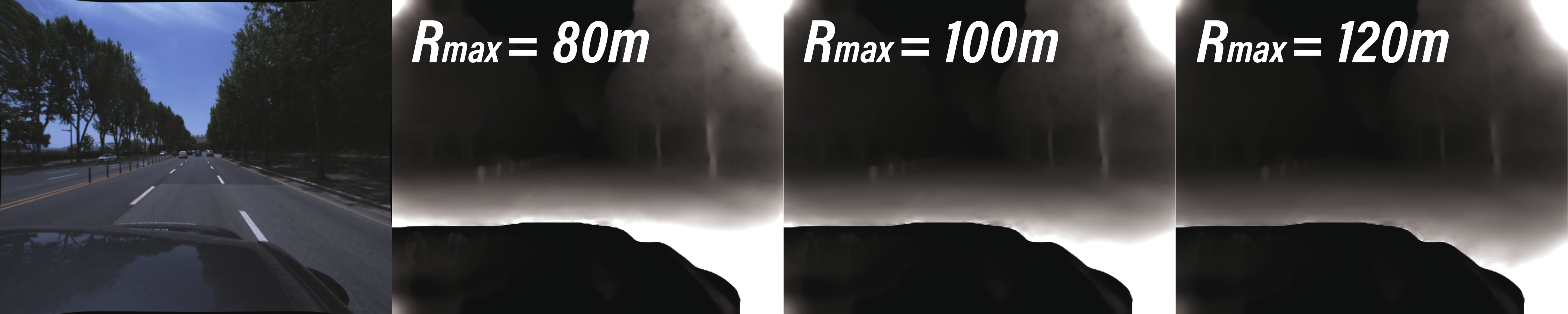}
	\caption
	{
		Depth augmentation during Phase 1. We multiplied a random scaling constant $r$ to the estimated depth values from the monocular disparity image, to randomly scale $\pm r$\% of the depth image. The examples of the depth modification are shown from the second to fourth columns.
	}
	\vspace{-8mm}
	\label{fig:aug}
\end{figure}

%% file: src/experimental_results.tex
\section{Experimental Results}
\label{sec:result}
\subsection{Place Recognition}
\subsubsection{Vision for Visibility Dataset}
\input{src/figtex/fig_pr.tex}
We trained and tested our algorithm on a public dataset ViViD++~\cite{lee2022vivid}, consisting of training set around $233$k images and $30$k scans before augmentation. The similarity constant $\tau$ is set to $0.5$, the margin $m$ is set to $0.1$ and the depth scaling variable $r$ is set to $20$. The sequences consisted of multiple repetitions over a similar trajectory during day and night, with changing lighting conditions. For the experiment, we set up a scenario of revisiting places at different times from prior visits. Because an image database could be constructed at the time when the lighting conditions were favorable, we used images from \textit{day2} as the database.

As illustrated in Figs.~\ref{fig:topn_recall} and~\ref{fig:precision_recall}, the re-identification of a place was easily accomplished with every module when matching images in similar lighting conditions. However, at night, image-based place recognition failed to identify the closest matches. Conversely, our cross-matching and point-cloud-based baseline were not affected by the lighting condition changes. Both the proposed method and the baseline use LiDAR scans, but our method does not require a point cloud database for matching. It should be noted that all the results in this subsection were obtained without using the information-based filtering presented in Section~\ref{sec:loop_method}. In Tables I, II, and III, we report the recall performance of every matching scenario within and across the data sources' modality, e.g. 3D-2D represents LiDAR queries matched into image database. As is evident from Tables~\ref{tab:top1_vivid} and ~\ref{tab:top1p_vivid}, matching between the identical modalities resulted in the best top-1 performance and our module performed better at 1\% recall. This is because of the limitations of cross-modal matching with unscaled depth images. For example, unconstrained depth images fail the matching. The cases are explained in more detail in Section~\ref{sec:failure}.
\input{src/figtex/table_vivid.tex}

\subsubsection{Oxford Robotcar Dataset}
To evaluate environmental changes other than appearance, we used another public dataset with
seasonal changes. The Oxford Robotcar Dataset~\cite{RobotCarDatasetIJRR} involves multiple runs over the same
trajectory, repeated forty-four times for more than a year. From the script that produces a 3D point cloud map from 2D scans using ground truth poses, we first constructed a point cloud map from the scans and transformed it
into a LiDAR range image using the projection method proposed in Section~\ref{sec:rangeimage}. The test and train
sequences were split by non-overlapping regions, 70\% and 30\% for training and testing, respectively, as 
in~\cite{cattaneo2020global}. The performance metrics of our module are listed in Table~\ref{tab:top1p_robotcar}. Our
module outperformed the baseline in cross-matching; however, showed lower performance when matching identical data types.
We presume that the lower performance was resulted from the limitations of the depth module and information loss
during the image-to-depth transformation in 2D-2D and the cropped LiDAR FoV in 3D-3D.

\input{src/figtex/table_oxford.tex}

\subsection{Loop Closure}
As mentioned in Section~\ref{sec:loop_method}, we filtered the raw correspondences from cross-matching with factor graph optimization  which was implemented using GTSAM~\cite{dellaert2012factor}. Fig.~\ref{fig:slam} presents the matching results for two different trajectories (\textit{day1} and \textit{night}). The raw correspondences from our module before filtering are shown in the left column, whereas the optimized correspondences after filtering are shown in the right column. Only matches with a descriptor distance lower than 0.1 were plotted, and only 46\% (285 out of 610) matches were correct; the others were false-positive loops. After information-based filtering, 90\% (87 out of 97) matches were found to be correct. There was a trade-off between the number of remaining matches and ratio of the correct match; the threshold $\tau$ applied to the trace of the information matrix should be raised if the locations from the geo-tags are unreliable.
\input{src/figtex/fig_pgo_rejection.tex}

\section{Discussions \& Future works}
\subsection{Image and Depth-Based Local Features}
\input{src/figtex/fig_match_daynight.tex}
\label{sec:depthmodules}
\input{src/figtex/fig_one_recall.tex}
To identify the image regions where the network extracts informative descriptors, the matching results of the local deep features are presented in Fig.~\ref{fig:localfeature}. The figure shows the matching results of the local descriptors for the RGB and depth images. The local descriptors were extracted from the outputs of their ground truth pairs from \texttt{conv5}. It can be seen that most image-based features are based on visual features such as crosswalks, color boundaries, or empty textures in the sky. Meanwhile, structural features like road signs, trees, or road boundaries are extracted in our encoder module. Assuming a LiDAR to camera matching case, these correspondences become a perspective-\textit{n}-point problem, and relative poses are solved by solutions such as EP\textit{n}P~\cite{lepetit2009epnp}.

\subsection{Image-Based Loop Closures for LiDAR-Based Systems}
\label{sec:one_recall}
In Fig.~\ref{fig:slam_lidar}, we compared the performance of our cross-matching with visual place recognition (VPR)-based loop closures after factor graph optimization. For the experiment, we generated noisy odometry with a heading error of 30$^\circ$ and assigned relative poses from top-1 estimates of every query image. For a fair comparison, the depth values were assigned from deskewed LiDAR scan for image-based features. Consequently, VPR-based loop closures succeeded only in the day-day matching but failed in night-day matching owing to the false-positive loop closures and the failure of depth estimator upon the illumination changes. However, our cross-matching based on the LiDAR scan query is nearly invariant to lighting conditions, and our module maintains a similar RMSE level on both matching scenarios.

\subsection{Limitations of Projections for Place Recognition}
\label{sec:failure}
\input{src/figtex/fig_failurecases.tex}
As mentioned in Section~\ref{sec:one_recall}, non-informative loop closures could be filtered out through pose graph optimization. However, some cases of false-positive loop closures still exist. Filtering based on the feature distance was insufficient for such failure cases, as shown in Fig.~\ref{fig:failure}. In the first case, close and unique objects are absent, and only distant landscapes are present, as shown in Fig.~\ref{fig:failure}(a). Large structures at a distance do not experience significant viewpoint changes after translation and the accurate location cannot be determined even when the matching score is high. As shown in the figure, large buildings do not experience viewpoint changes, although the lower picture of Fig.~\ref{fig:failure}(a) were taken 86 m apart from the upper one. Moreover, urban structures such as street lamp poles are non-informative and highly repetitive, thus lowering the feature distances of false positive matches. For cases with underdetermined poses, there were no sufficient planes to constrain the translation along a specific axis, as shown in Fig.~\ref{fig:failure}(b). These circumstances were the main sources of false-positive loop closures in our experiments.

%% file: src/figtex/fig_pr.tex
\begin{figure}[!b]
	\captionsetup{font=footnotesize}
	\vspace{-4mm}
	\centering
	\subfigure{%
		\includegraphics[width=0.6\columnwidth]{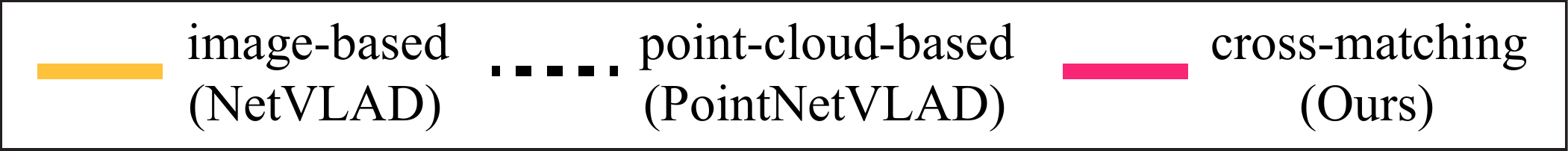}
	}\vspace{-1mm}
	
	\captionsetup{font=footnotesize}
	\subfigure{%
		\includegraphics[width=0.33\columnwidth]{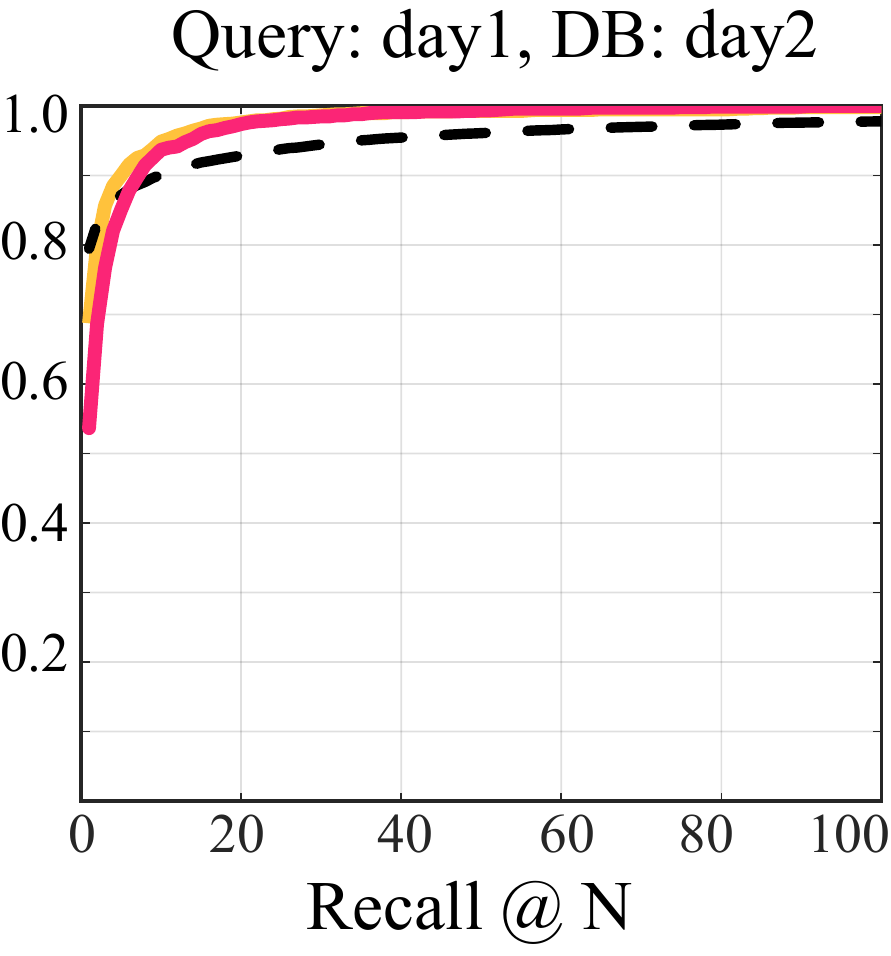}
	}
	\subfigure{%
		\includegraphics[width=0.33\columnwidth]{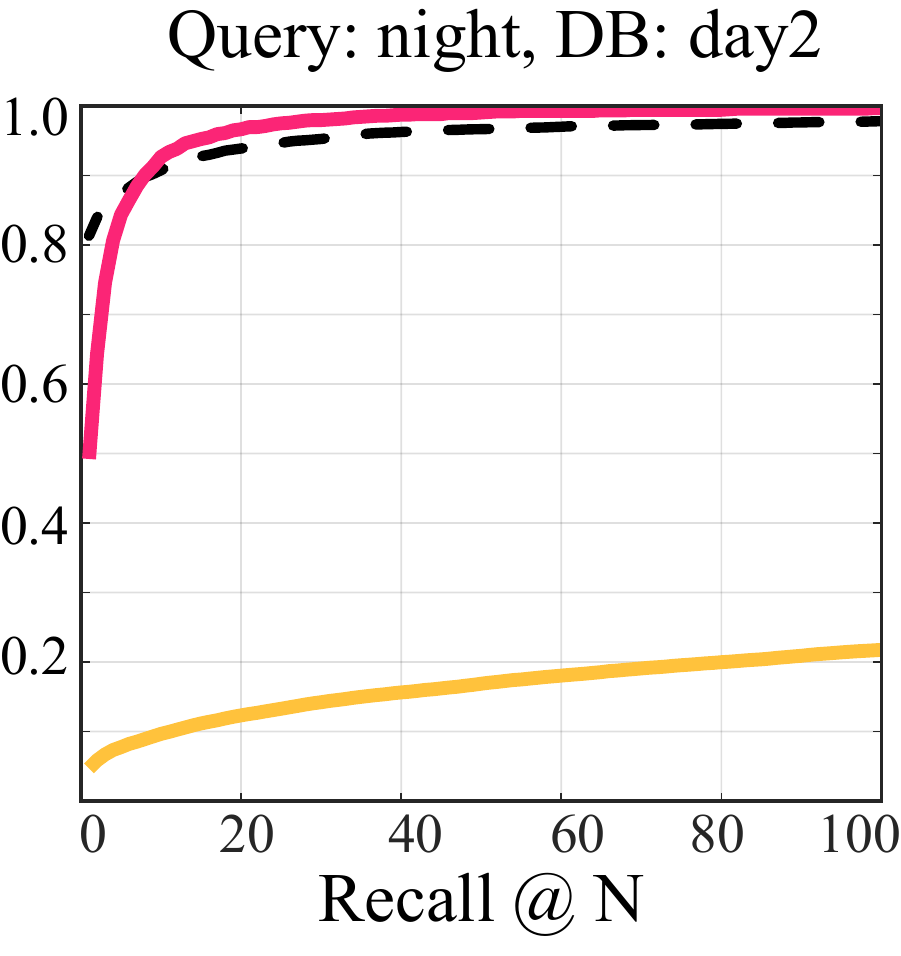}
	}\vspace{-2mm}
	\caption
	{
		Top N recall of visual place recognition (solid yellow), point-cloud-based place recognition (dotted black), and the proposed cross-modal place recognition (magenta solid).
	}
	\label{fig:topn_recall}
	
	\centering
	\subfigure{%
		\includegraphics[width=0.55\columnwidth]{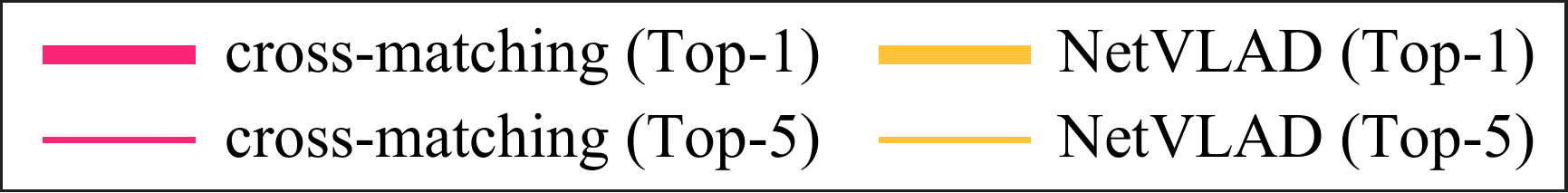}
	}\vspace{-2mm}
	
	\captionsetup{font=footnotesize}
	\subfigure{%
		\includegraphics[width=0.36\columnwidth]{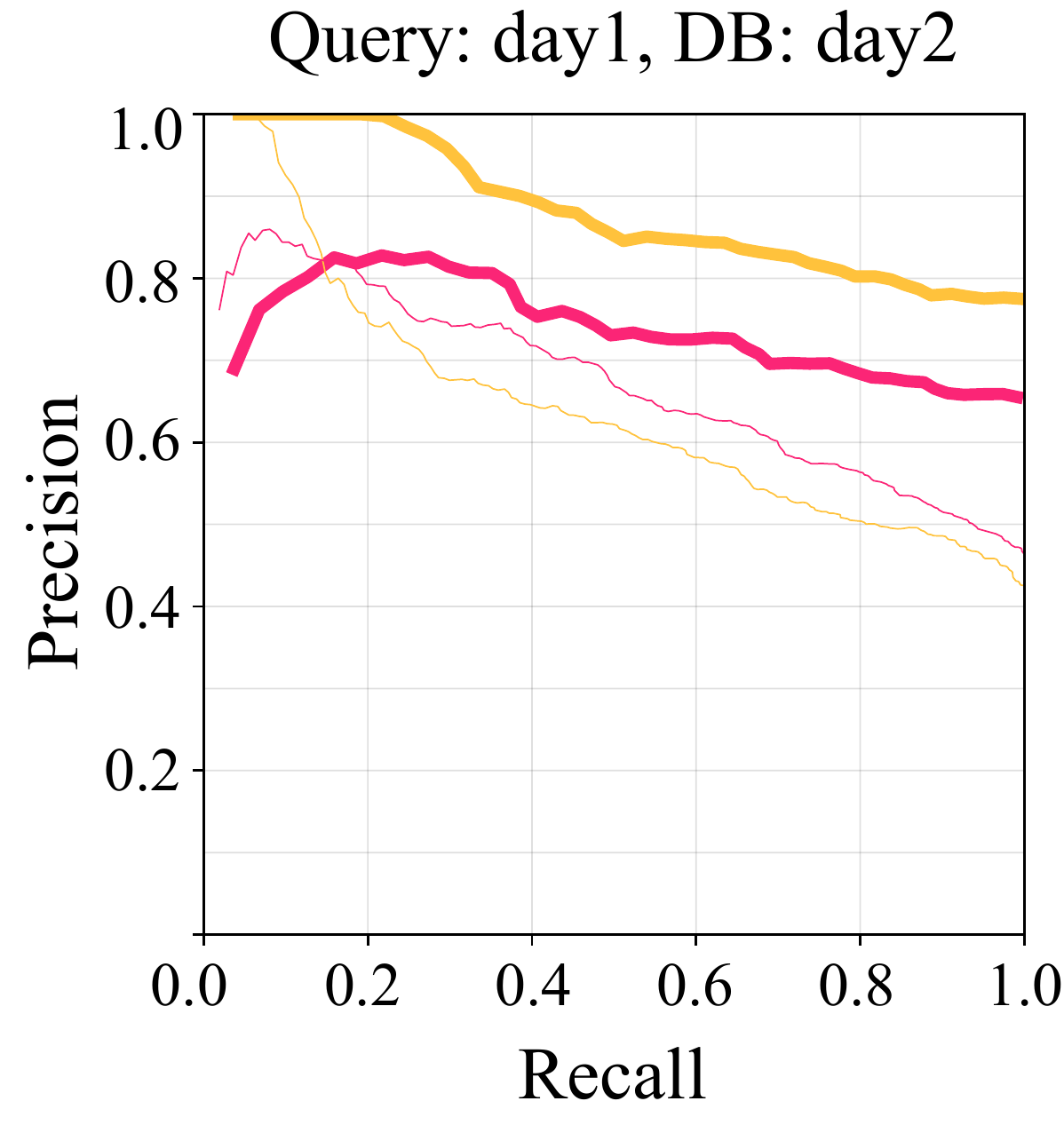}
	}
	\subfigure{%
		\includegraphics[width=0.36\columnwidth]{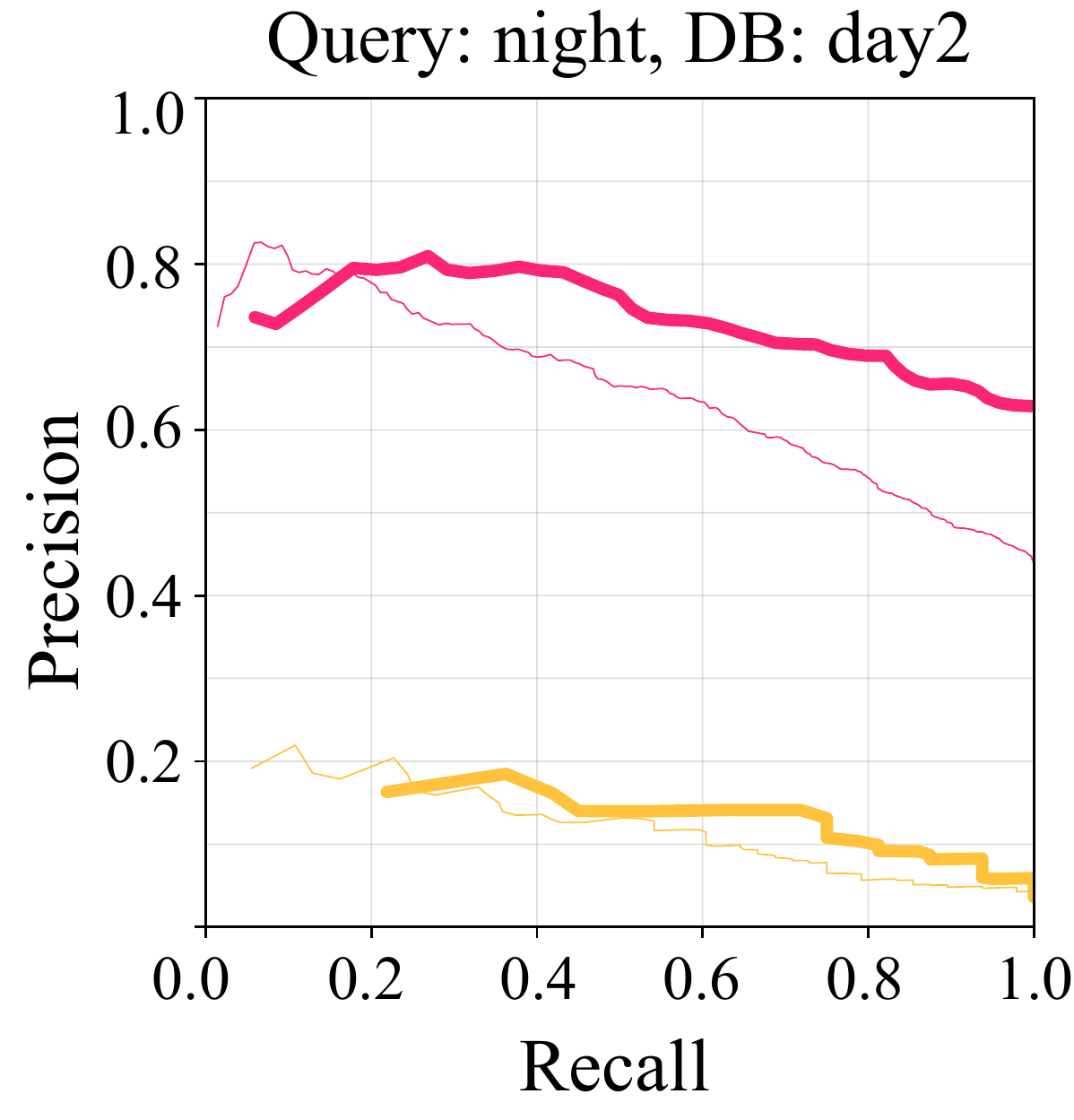}
	}\vspace{-3mm}
	\caption
	{
		Precision-recall curve of visual place recognition (yellow) and the proposed cross-modal place recognition (magenta).
	}
	\label{fig:precision_recall}
	
\end{figure}

%% file: src/figtex/table_vivid.tex
\begin{table}[t]
	\captionsetup{font=footnotesize}
	\renewcommand{\arraystretch}{1.2}
	\centering
	\setlength{\tabcolsep}{8pt}
	\caption{\sc Recalls of day1-day2 matching.}
	\begin{tabular}{|c|c|c|c|c|}
		\hline
		\textbf{Top-1} & 2D-2D & 3D-3D & 2D-3D & 3D-2D \\ \hline
		PointNetVLAD~\cite{uy2018pointnetvlad} & -     & \textbf{0.7946}& -     & -     \\ \hline
		NetVLAD~\cite{arandjelovic2016netvlad} & 0.7016& -     & -     & -     \\ \hline
		GCL~\cite{leyvavallina2021gcl}         & \textbf{0.8205}& -     & -     & -     \\ \hline
		Ours                                   & 0.6921& 0.5806& \textbf{0.6087}& \textbf{0.5177}\\ \hline \hline
		\textbf{Top-1\%}& 2D-2D & 3D-3D & 2D-3D & 3D-2D \\ \hline
		PointNetVLAD~\cite{uy2018pointnetvlad} & -     & 0.9124& -     & -     \\ \hline
		NetVLAD~\cite{arandjelovic2016netvlad} & 0.9252& -     & -     & -     \\ \hline
		GCL~\cite{leyvavallina2021gcl}         & 0.9606& -     & -     & -     \\ \hline
		Ours                                   & \textbf{0.9685}& \textbf{0.9606}& \textbf{0.9598}& \textbf{0.9457}\\ \hline
	\end{tabular}
	\label{tab:top1_vivid}
\end{table}

\begin{table}[t]
	\captionsetup{font=footnotesize}
	\renewcommand{\arraystretch}{1.2}
	\centering
	\setlength{\tabcolsep}{8pt}
	\caption{\sc Recalls of night-day2 matching.}
	\begin{tabular}{|c|c|c|c|c|}
		\hline
		\textbf{Top-1} & 2D-2D & 3D-3D & 2D-3D & 3D-2D \\ \hline
		PointNetVLAD~\cite{uy2018pointnetvlad} & -     & \textbf{0.8133}& -     & -     \\ \hline
		NetVLAD~\cite{arandjelovic2016netvlad} & 0.0093& -     & -     & -     \\ \hline
		GCL~\cite{leyvavallina2021gcl}         & 0.0335& -     & -     & -     \\ \hline
		Ours         & 0.0078& 0.5482& 0.0046& \textbf{0.4938}\\ \hline \hline
		\textbf{Top-1\%}& 2D-2D & 3D-3D & 2D-3D & 3D-2D \\ \hline
		PointNetVLAD~\cite{uy2018pointnetvlad} & -     & 0.9222& -     & -     \\ \hline
		NetVLAD~\cite{arandjelovic2016netvlad} & 0.1626& -     & -     & -     \\ \hline
		GCL~\cite{leyvavallina2021gcl}         & 0.1439& -     & -     & -     \\ \hline
		Ours         & 0.0553& \textbf{0.9487}& 0.0529& \textbf{0.9339}\\ \hline
	\end{tabular}
	\label{tab:top1p_vivid}
	\vspace{-4mm}
\end{table}

%% file: src/figtex/table_oxford.tex
\begin{table}[b]
	\captionsetup{font=footnotesize}
	\vspace{-2mm}
	\renewcommand{\arraystretch}{1.2}
	\centering
	\setlength{\tabcolsep}{8pt}
	\caption{\sc{Top-1\% Recalls from Oxford robotcar dataset.}}
	\begin{tabular}{c|c|c|c|c|}
		\cline{2-5}
		& 2D-2D & 3D-3D & 2D-3D & 3D-2D \\ \hline
		\multicolumn{1}{|c|}{Cattaneo \textit{et al}.~\cite{cattaneo2020global}} & \textbf{0.9663} & \textbf{0.9843} & 0.7728 & 0.7044 \\ \hline
		\multicolumn{1}{|c|}{Ours}    & 0.8414 & 0.8298 & \textbf{0.8123} &  \textbf{0.7384} \\ \hline
	\end{tabular}
	\label{tab:top1p_robotcar}
\end{table}

%% file: src/figtex/fig_pgo_rejection.tex
\begin{figure}[!h]
	\centering
	\captionsetup{font=footnotesize}
	\includegraphics[width=\columnwidth]{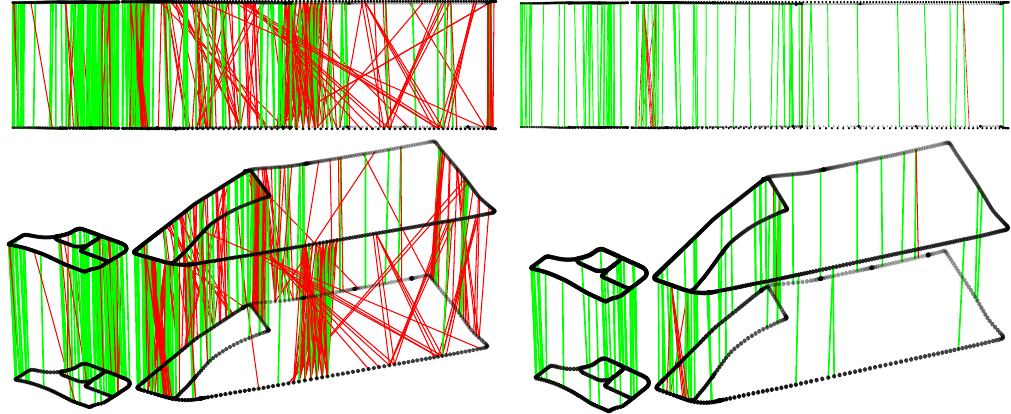}
	\caption
	{
		Raw (left) and filtered (right) loop closures based on feature distance and information. True-positive loop closures are indicated in green and false-positives in red.
	}
	\label{fig:slam}
\end{figure}

%% file: src/figtex/fig_match_daynight.tex
\begin{figure}[!h]
	\centering
	\vspace{-2mm}
	\captionsetup{font=footnotesize}
	\subfigure[]{%
		\includegraphics[width=0.48\columnwidth]{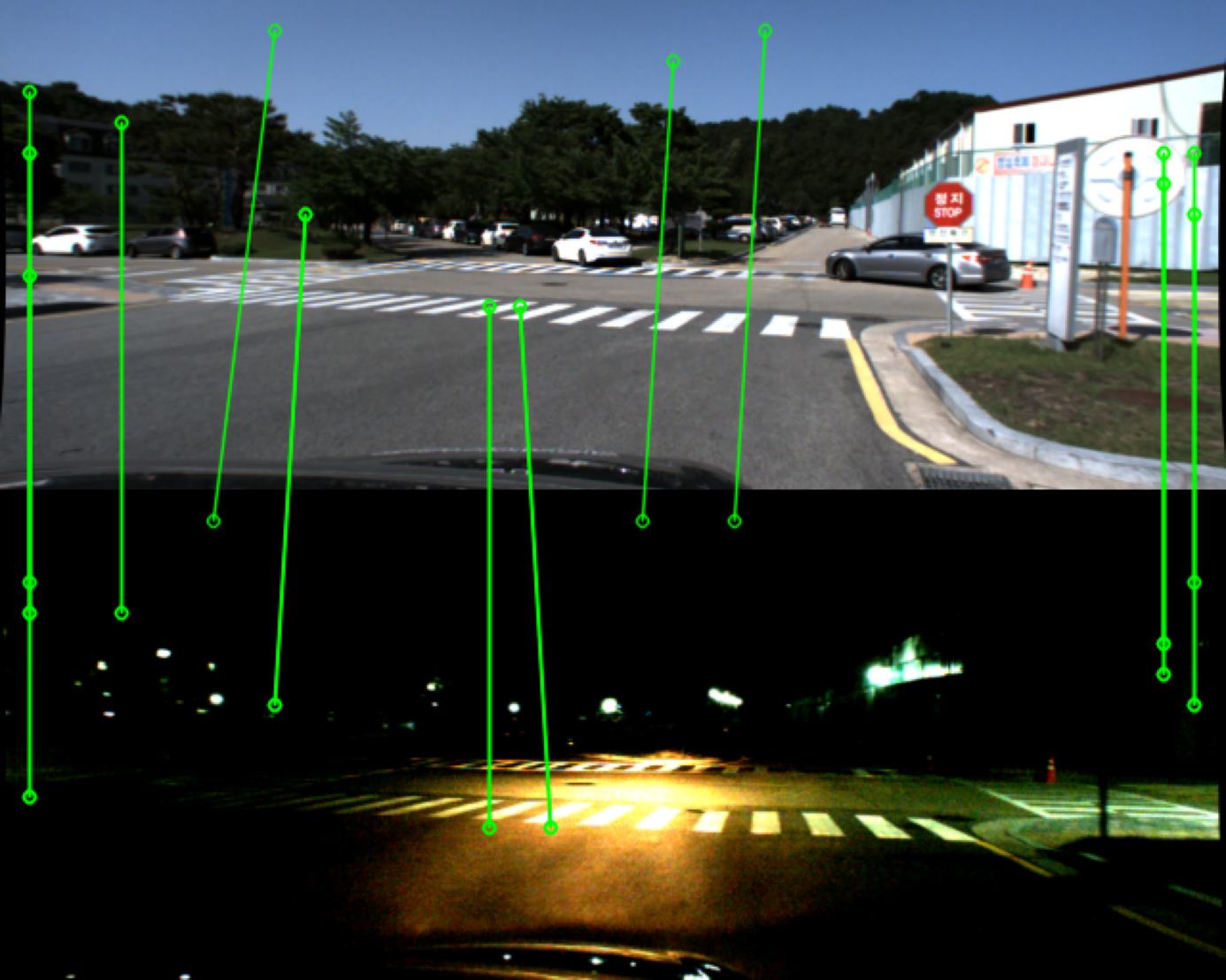}
	}%
	\subfigure[]{%
		\includegraphics[width=0.48\columnwidth]{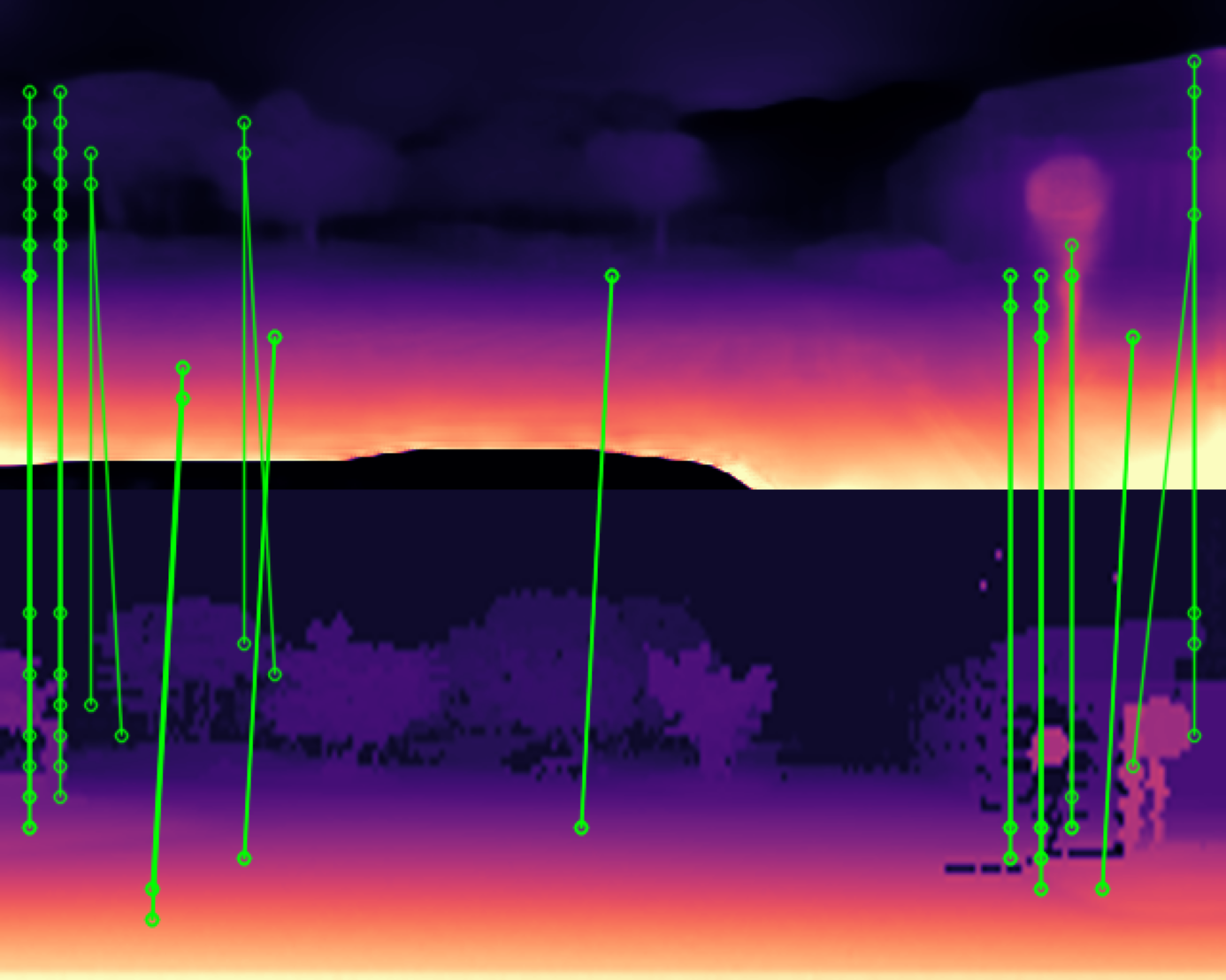}
	}
	\vspace{-2mm}
	\caption
	{
		Local feature matches of the same place at different times, from (a) image-based features (query at night and database from daytime) and (b) depth-based local features (LiDAR query at night and database of disparity image, transformed from daytime). In contrast to few and inaccurate match pairs (e.g. matches in the featureless sky) in the RGB images, the depth images are invariant and most features are re-identified despite the appearance change.
	}
	\vspace{-2mm}
	\label{fig:localfeature}
	
\end{figure}

%% file: src/figtex/fig_one_recall.tex
\begin{figure*}[t]
	\centering
	\captionsetup{font=footnotesize}
	\vspace{-2mm}
	\includegraphics[width=\textwidth]{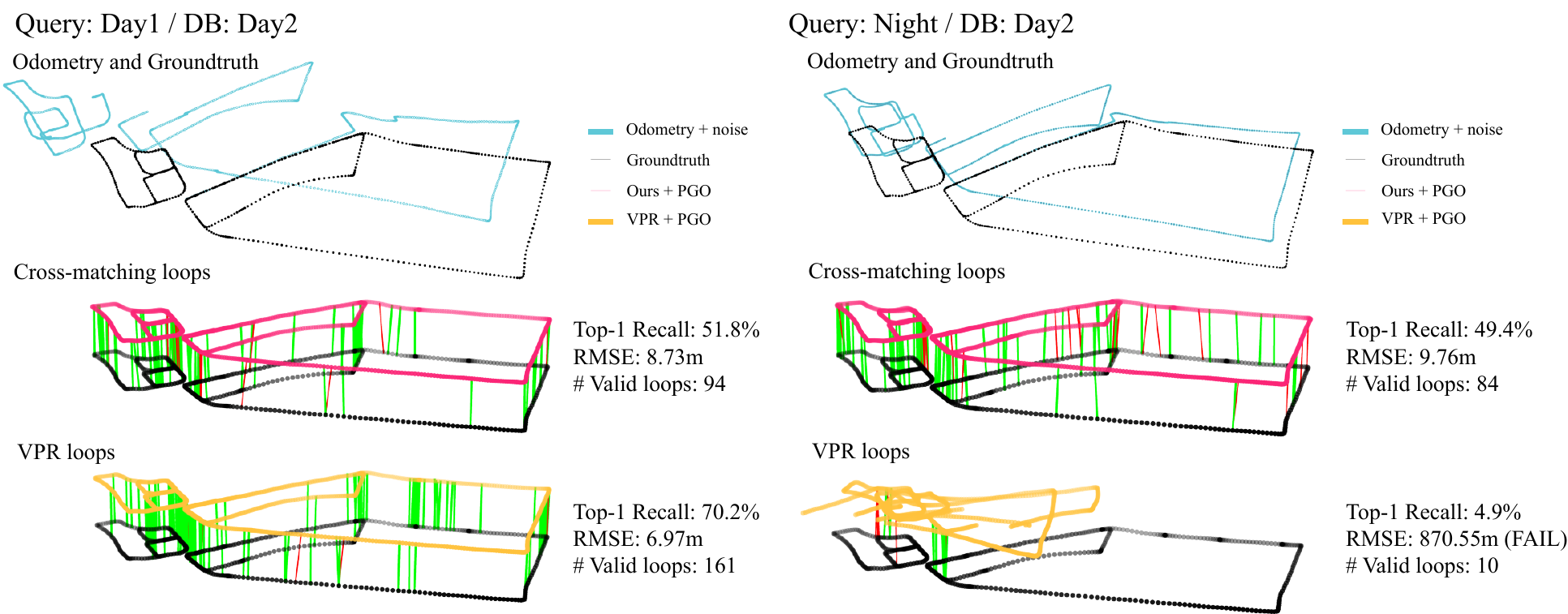}
	\caption
	{
		Odometry and trajectory after optimization with loop closure. We first constructed relative poses from LiDAR odometry~\cite{lin2019loam_livox} and added noise to both the relative and starting poses. The random noise assigned to each relative pose was sampled from the uniform distribution. We then obtained the optimized pose graph by composing a pose graph and solving it with an LM optimizer. Both our module and VPR-based loops succeeded in reducing the noise from the odometry. However, only our module based on cross-matching could converge at night-to-day matching sequence.
	}
	\label{fig:slam_lidar}
\end{figure*}

%% file: src/figtex/fig_failurecases.tex
\begin{figure}[b]
	\centering
	\vspace{-4mm}
	\captionsetup{font=footnotesize}
	\subfigure[]{
		\vspace{-2mm}
		\includegraphics[width=0.45\columnwidth]{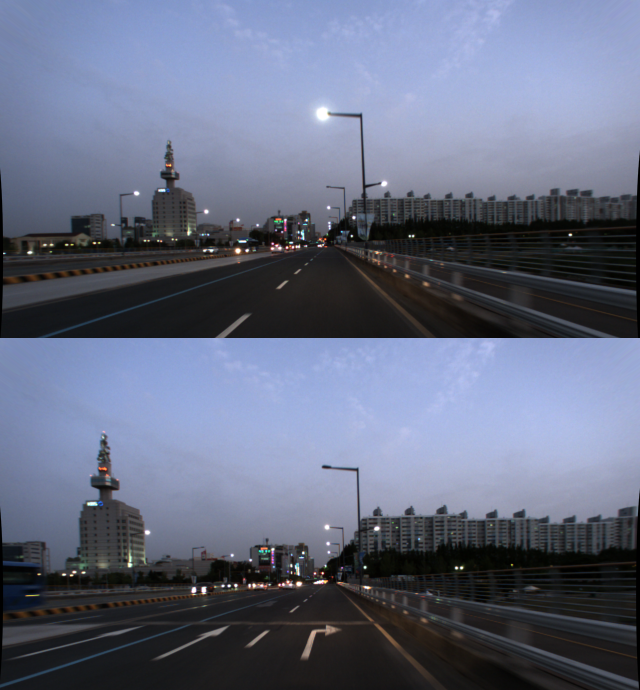}
	}%
	\subfigure[]{
		\vspace{-2mm}
		\includegraphics[width=0.45\columnwidth]{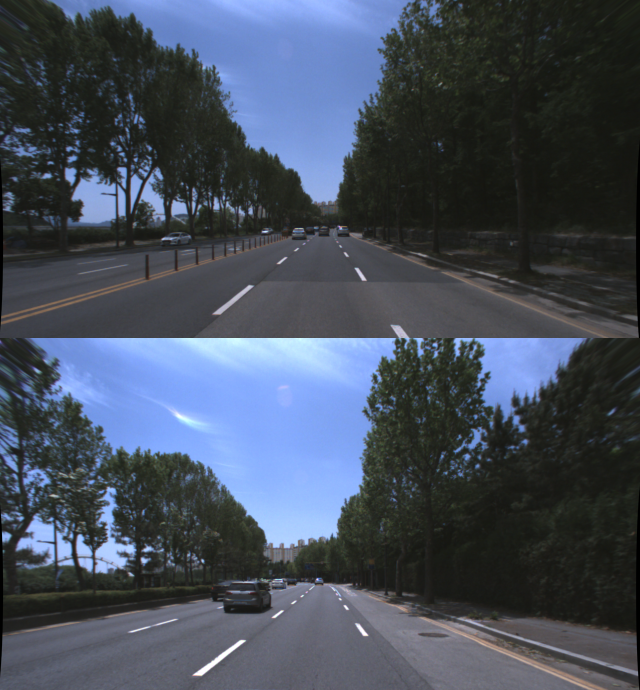}
	}
	\vspace{-3mm}
	\caption
	{
		Common failure cases of our perspective depth place recognition. Scenes with ambiguous relative depths failed. For example, when (a) major landscape at far away makes triangulation unreliable, or (b) translation through an axis becomes unconstrained by the corridor-like environment.
	}
	\label{fig:failure}
	\vspace{-3mm}
	
\end{figure}

%% file: src/conclusion.tex
\section{Conclusion}
\label{sec:conclusion}

In this paper, we proposed a cross-modal place recognition algorithm for LiDAR-camera systems. We experimentally demonstrated that images and point clouds can be transformed into shared neural embedding. The developed method is expected to widen the possibility of fusion between two complementary sensors, namely LiDAR and camera, starting from data intercompatibility and further to spatial AI.